\documentclass[letterpaper, 10 pt, conference]{ieeeconf} 

\IEEEoverridecommandlockouts
\usepackage{cite}
\usepackage{amsmath,amssymb,amsfonts}
\usepackage{algorithm}
\usepackage{algpseudocode}
\usepackage{graphicx}
\usepackage{textcomp}
\usepackage{xcolor}
\usepackage{booktabs}
\usepackage{multirow}
\usepackage{arydshln} 
\usepackage{hyperref}

\usepackage{cuted}
\usepackage{capt-of} 

\def\BibTeX{{\rm B\kern-.05em{\sc i\kern-.025em b}\kern-.08em
    T\kern-.1667em\lower.7ex\hbox{E}\kern-.125emX}}
    
\newcommand{\R}{\mathbb{R}}

\newcommand{\method}{\emph{PriMoRetarget}}

\title{\LARGE \bf
PriMoRetarget: Humanoid Motion Transfer Through Prioritized Motion Retargeting
}

\author{
Nicholas Ioannidis$^{\ast1}$,
Dun Yang$^{\ast 1}$,
Max Smith$^{2}$,
\\
Tue Vu$^{2}$,
Ziyu Zhang$^{1}$,
Ali Agha$^{2}$,
Shayegan Omidshafiei$^{2}$,
Xue Bin Peng$^{1,3,4}$%
\\[3pt]
$^{1}$Simon Fraser University,
$^{2}$FieldAI,
$^{3}$NVIDIA,
$^{4}$Alberta Machine Intelligence Institute
\\[5pt]
Project page: \href{https://nickioan.github.io/primoretarget-page/}{\nolinkurl{primoretarget-page}}
\thanks{$^{\ast}$Joint First Authors}
}

\begin{document}
\maketitle
\thispagestyle{empty}
\pagestyle{empty}

\begin{strip}
  \centering
  \includegraphics[width=\textwidth]{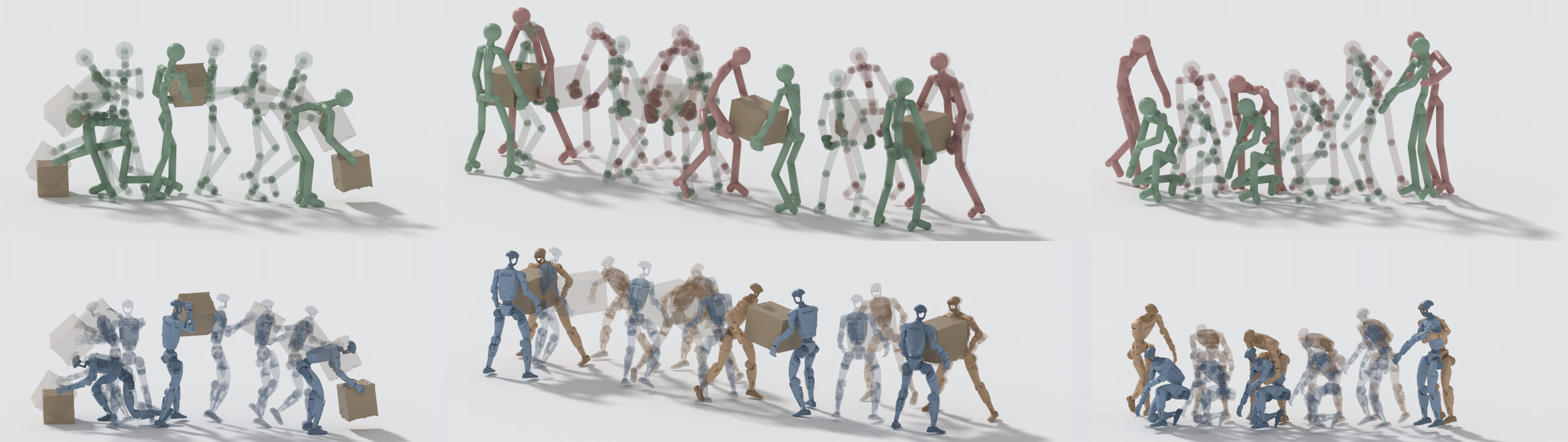}
   \captionof{figure}{
   \textbf{\method~is applied to motions depicting different interactions.}
  \textbf{Top:} Source motion capture clip. \textbf{Bottom:} Motions retargeted to Unitree G1 for character--object manipulation, collaborative object manipulation, and character--character assistance.}
  \label{fig:teaser}
\end{strip}

\begin{abstract}
Motion retargeting is a critical preprocessing step for transferring human motion data to humanoid robots. 
While existing methods are effective for single-character motions, interactions involving multiple characters and objects introduce additional contacts and spatial relationships that must be preserved. 
Retargeting is commonly formulated as a weighted multi-objective optimization problem, where conflicting objectives can require careful tuning of relative weights and can still compromise important motion characteristics. 
We introduce \method, a prioritized motion-retargeting framework that replaces relative objective weighting with an explicit priority hierarchy.
Lower-priority updates are constrained to preserve higher-priority objectives, allowing critical characteristics such as foot placement and interaction contacts to take precedence over lower-priority pose objectives. 
The method is evaluated on four datasets spanning single-character motion, character--object manipulation, and character--character interactions. 
The resulting motions better preserve task-relevant interactions and provide reference trajectories that can be tracked more reliably than those produced by widely used retargeting methods.
\end{abstract}

\section{Introduction}

For humanoid robots, human motion-capture data provides a rich source of reference behaviors despite morphological differences between humans and robots. 
Motion retargeting bridges this embodiment gap by transferring motion across embodiments while preserving the defining characteristics of the source motion.
A long line of prior work has studied retargeting single-character motions~\cite{gleicher1998retargetting,araujo2025retargeting,muller2026reactor}. 
However, motions involving interactions with objects or other characters remain substantially more challenging, as successful retargeting must preserve not only individual body poses but also the contacts, spatial relationships, and coordinated movements that define the interaction. 
Errors in these relationships can produce physically infeasible reference motions and increase the burden on downstream learning and control.

Many commonly used retargeting methods formulate these requirements as a multi-objective optimization problem, where each objective captures a characteristic of the source motion to preserve in the target motion~\cite{gleicher1998retargetting,ho2010spatial,araujo2025retargeting,yang2025omniretarget}. 
Such objectives may include matching source keypoints, preserving foot placement and interaction contacts, and respecting the robot's joint limits. 
Jointly optimizing these objectives is nontrivial because they can conflict: improving pose fidelity, for example, may degrade contact preservation or foot placement. 
These conflicts become increasingly pronounced as additional characters and objects introduce more contacts and spatial relationships that must be preserved.
To resolve these conflicts, retargeting methods typically rely on carefully tuned objective weights that reflect task- and morphology-specific priorities~\cite{gleicher1998retargetting,ho2010spatial,araujo2025retargeting,yang2025omniretarget}.
However, weighted objectives encode only relative preferences: improving a lower-weighted objective may still come at the expense of a higher-weighted objective.
Consequently, these weights do not guarantee that desired priority relationships between objectives are preserved.
Achieving the desired priority relationships therefore often requires careful manual tuning of the objective weights.

Two broad classes of methods address conflicting objectives without weight tuning: 
(1) prioritized optimization methods, which organize objectives into a hierarchy~\cite{nakamura1987task,siciliano1991general,escande2014hierarchical,de2010feature}, and 
(2) gradient-surgery methods, which modify interfering gradients across objectives~\cite{yu2020pcgrad,liu2021cagrad}.
Our approach draws on both, using an explicit priority hierarchy to specify which objectives should take precedence and gradient projection to enforce these priorities during optimization.

In this work, we propose \method, a prioritized motion-retargeting framework that replaces soft objective weighting with an explicit priority hierarchy.
Lower-priority updates are restricted to preserve higher-priority objectives, allowing critical requirements to take precedence over lower-priority preferences.
We show that our method is compatible with commonly used motion-retargeting objectives and apply our framework to four datasets spanning multi-character and object interactions. We evaluate the resulting motions using a range of kinematic metrics and assess their effectiveness as reference trajectories for training reinforcement learning (RL) policies. 
Our method substantially reduces common retargeting artifacts that compromise physical feasibility, including foot sliding and contact errors. The code and retargeted motion data will be publicly released upon publication.

\section{Related Work}

\paragraph{Motion retargeting}
Motion retargeting is commonly posed as a constrained optimization problem~\cite{gleicher1998retargetting}, often formulated as inverse kinematics (IK) to reconstruct source-motion features such as joint positions, orientations, and end-effector locations~\cite{gleicher1998retargetting,luo2023perpetual,araujo2025retargeting,muller2026reactor}.
While most methods assume similar source and target morphologies, prior work has also considered retargeting across different body structures, such as retargeting from humanoids to quadrupeds~\cite{yamane2010animating,seol2013creature}.

Humans rarely act in isolation; instead, human motions often involve interactions with objects, terrain, and other actors.
Retargeting such interaction-rich motions therefore requires preserving not only individual poses, but also the relationships that define these interactions.
Early work represented interactions through relative spatial geometry~\cite{ho2010spatial}, with subsequent methods applying similar representations to bipedal robots~\cite{nakaoka2012interaction}, terrain and object interactions~\cite{yang2025omniretarget}, and multi-character interactions~\cite{zhang2023simulation,huang2026beyond}. 
Trajectory optimization and simulation have also been incorporated into retargeting pipelines to improve dynamic feasibility~\cite{wang2026spark,zhang2026kinodynamic,dhedin2026dynaretarget, muller2026reactor}, though these methods typically rely on IK-based retargeting methods to provide a kinematic reference trajectory.

Despite their differences in their representation and optimization procedure, many modern retargeting methods formulate motion retargeting as a weighted multi-objective optimization problem.
When different objectives conflict, their relative importance is specified through weights.
These trade-offs can be difficult to tune and may need to be adjusted for different motions and target morphologies.
Our work alleviates the need for manual weight tuning by organizing multiple objective functions into a prioritized hierarchy, allowing higher-priority physical requirements to take precedence over lower-priority preferences.

\paragraph{Data-driven retargeting}
As an alternative to optimization-based methods, data-driven retargeting methods learn mappings between different morphologies directly from motion data, without requiring explicitly defined motion retargeting objectives~\cite{villegas2018neural,lim2019pmnet,aberman2020skeleton,lee2023same,hu2023pose,zhang2024unified}. 
By learning representations that disentangle motion from embodiment, these methods can support retargeting across substantially different skeletal structures~\cite{aberman2020skeleton,lee2023same,hu2023pose,zhang2024unified}.
Contact-aware methods further incorporate geometric and contact information to improve the physical consistency of the retargeted motion~\cite{villegas2021contact,cheynel2025reconform}.

While data-driven retargeting can be effective, these methods are generally constrained by the motions and morphologies represented in their training data and may struggle to generalize beyond this distribution.
In contrast, optimization-based methods do not require motion- or morphology-specific training data and can therefore be applied more broadly across behaviors and embodiments. 
As a result, optimization-based retargeting has thus far seen wider practical adoption than its data-driven counterparts.

\paragraph{Multi-objective optimization}
Balancing competing objectives is a central challenge in multi-objective optimization. 
Adaptive reweighting methods adjust loss weights during optimization~\cite{kendall2018multi,chen2018gradnorm}, while CAGrad accounts for worst-case objective performance~\cite{liu2021cagrad}. 
Gradient-surgery methods such as PCGrad~\cite{yu2020pcgrad} instead modify gradients to reduce interference. 
However, these methods generally do not enforce an ordering of importance. 
In motion retargeting, some objectives may require strict precedence; for example, preserving a foot contact may be high-priority than matching upper-body pose.
Prioritized optimization encodes such precedence through an explicit objective hierarchy.

Prioritized optimization has been widely used in robotics to exploit kinematic redundancy. 
Higher-priority objectives often admit multiple solutions, leaving residual degrees of freedom available for lower-priority objectives.
Task-priority methods exploit this redundancy by restricting lower-priority optimization to solutions that preserve higher-priority objectives~\cite{liegeois1977automatic,nakamura1987task,siciliano1991general}.
This principle has been applied to IK for articulated characters~\cite{baerlocher2004inverse} and whole-body humanoid control~\cite{sentis2005synthesis,de2010feature,kanoun2011kinematic,escande2014hierarchical,dietrich2015overview}. 
Prior work has primarily considered online control and model-predictive settings.
In contrast, we show that prioritized optimization is also effective for motion retargeting and introduce techniques that enable its efficient application to long motion sequences, multiple interacting characters, and general non-quadratic retargeting objectives.

\section{Prioritized Motion Retargeting}
\label{sec:method}
Motion retargeting is often formulated as an optimization problem~\cite{gleicher1998retargetting}, in which the objective is defined as a weighted sum of multiple loss terms:
\begin{equation}
x^\star = \arg\min_x  \sum_{k=1}^{K} w^k L^k(x),
\label{eq:weighted-retargeting}
\end{equation}
where $x$ denotes the target motion, $L^k$ is a loss function, and $w^k$ determines its relative contribution. This formulation provides a flexible way to combine different objectives, but 
when objectives conflict, improving one may come at the expense of another, even when one represents a more important requirement.

To resolve conflicts between different objectives, we propose explicitly ordering the objectives into a prioritized hierarchy. Let ${L^1,\ldots,L^K}$ denote objectives ordered from highest to lowest priority and let $S^0$ be the set of all target motions. At priority level $k$, objective $L^k$ is minimized over the solution set $S^{k-1}$ of motions that remain optimal for all higher-priority objectives, i.e., over motions that remain optimal for all higher-priority objectives,
\begin{equation}
x_k^\star \in \arg\min_{x\in S^{k-1}} L^k(x),
\label{eq:priority-level-optimum}
\end{equation}
and the corresponding solution set becomes
\begin{equation}
S^k =
\left\{
x\in S^{k-1}
\;\middle|\;
L^k(x)=L^k(x_k^\star)
\right\}.
\label{eq:prioritized-solution-set}
\end{equation}
Although the recursive formulation defines the hierarchy sequentially, we enforce it locally at each optimization step by restricting lower-priority updates to directions that preserve higher-priority objectives to first order. 
We next describe this enforcement and an efficient implementation for long sequences and interacting actors.

\subsection{Optimizing in the Nullspace}

\begin{figure}[!tb]
\centering
\includegraphics[width=0.85\columnwidth]{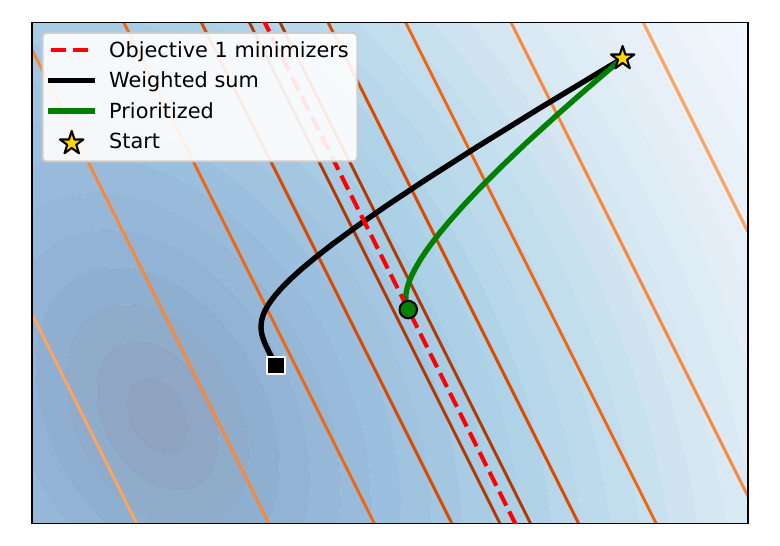}\\
\vspace{-0.3cm}
\caption{Weighted versus prioritized optimization for two competing objectives. The weighted solution compromises between objectives, while prioritized optimization restricts lower-priority updates to directions that preserve the optimal solution for Objective~1.}
\label{fig:toy_example}
\vspace{-0.3cm}
\end{figure}

Consider a higher-priority objective $L^{\mathrm{H}}$ and a lower-priority objective $L^{\mathrm{L}}$. Let
$f^{\mathrm{H}}(x)\in\mathbb{R}^{m}$ denote the vector-valued error of the higher-priority objective, with
$L^{\mathrm{H}}(x)=\|f^{\mathrm{H}}(x)\|^2$. For example, $f^{\mathrm{H}}$ may represent the 3D position errors of several corresponding bodies. We seek an update $\delta x$ that improves $L^{\mathrm{L}}$ without changing the higher-priority error,
\begin{equation}
f^{\mathrm{H}}(x+\delta x)=f^{\mathrm{H}}(x).
\label{eq:priority-error-preservation}
\end{equation}
Preserving the error vector, rather than only its scalar loss, prevents compensation across error components.

For a small update $\delta x$, we linearize the higher-priority error around the current parameters,
\begin{equation}
f^{\mathrm{H}}(x+\delta x)
\approx
f^{\mathrm{H}}(x)+J^{\mathrm{H}}\delta x,
\label{eq:priority-linearization}
\end{equation}
where $J^{\mathrm{H}}=\partial f^{\mathrm{H}}/\partial x$. Combining Eqs.~\ref{eq:priority-error-preservation} and~\ref{eq:priority-linearization} gives $J^{\mathrm{H}}\delta x=0$. Updates $\delta x$ that satisfy this relationship lie in the \emph{nullspace} of $J^{\mathrm{H}}$ and are directions that preserve the higher-priority error to first order~\cite{liegeois1977automatic,de2010feature,baerlocher2004inverse}.

$\delta x$ is obtained from gradient-based updates. 
Let $g^{\mathrm{L}}=\nabla_x L^{\mathrm{L}}$ denote the lower-priority gradient. 
We project $g^{\mathrm{L}}$ into the nullspace of $J^{\mathrm{H}}$ using the nullspace projector~\cite{baerlocher2004inverse},
\begin{equation}
P^{\mathrm{H}}
=
I-(J^{\mathrm{H}})^\top
\left(
J^{\mathrm{H}}(J^{\mathrm{H}})^\top
\right)^{-1}
J^{\mathrm{H}},
\label{eq:priority-projector}
\end{equation}
yielding the projected update
\begin{equation}
x \gets x-\alpha P^{\mathrm{H}}g^{\mathrm{L}}.
\label{eq:priority-projected-update}
\end{equation}
Figure~\ref{fig:toy_example} illustrates this projection for two competing quadratic objectives. Unlike a weighted sum, which converges to a compromise between the objectives, the projected lower-priority update is restricted to directions that preserve the higher-priority error to first order.

In practice, $J^{\mathrm{H}}$ may be rank deficient or poorly conditioned, so we use the damped projector
\begin{equation}
P^{\mathrm{H},\lambda}
=
I-(J^{\mathrm{H}})^\top
\left(
J^{\mathrm{H}}(J^{\mathrm{H}})^\top+\lambda I
\right)^{-1}
J^{\mathrm{H}},
\label{eq:priority-damped-projector}
\end{equation}
where $\lambda>0$ is a small damping factor. 

\subsection{Efficient Prioritized Optimization}
\label{sec:efficient}
Applying prioritized optimization to full motion sequences introduces challenges in computational cost, objective placement, and initialization.

\paragraph{Per-frame separability}
Computing the projection over an entire motion sequence is expensive, as the dense projection matrix grows quadratically with the number of frames. 
Many retargeting objectives, however, are separable across frames: the error vector at each frame depends only on that frame’s motion parameters. 
We exploit this structure by constructing the prioritized hierarchy from per-frame objectives, yielding Jacobians that are block diagonal in time.

At priority level $k$, let $J^{<k}_t$ denote the Jacobian of all higher-priority objectives at frame $t$, obtained by stacking their individual Jacobians,
\begin{equation}
J^{<k}_t =
\begin{bmatrix}
(J^1_t)^\top & \cdots & (J^{k-1}_t)^\top
\end{bmatrix}^{\top}.
\label{eq:priority-stacked-jacobian}
\end{equation}
Let $m$ denote the number of higher-priority error components per frame. 
Over $T$ frames, the full Jacobian is block diagonal in time, so the $(Tm)\times(Tm)$ projection decomposes into $T$ independent per-frame systems. 
For the gradient $g^k_t$ of objective $k$ at frame $t$, the projected gradient is
\begin{equation}
\tilde{g}^k_t = \left(I -
(J_t^{<k})^\top
\left(
J_t^{<k}(J_t^{<k})^\top+\lambda I
\right)^{-1}
J_t^{<k}\right)g^k_t.
\label{eq:priority-per-frame-projection}
\end{equation}
The $T$ systems are evaluated in parallel as a batched operation. Each involves only an $m\times m$ linear system, reducing the computational complexity from $O(T^3m^3)$ to $O(Tm^3)$. 

\paragraph{Free objectives}
Some objectives, such as temporal smoothness and other regularizers, do not fit conveniently within a strict priority hierarchy. These objectives often act over many or all degrees of freedom. Assigning them a high priority can therefore leave little or no degrees of freedom in the nullspace for lower-priority objectives, while assigning them a low priority can substantially restrict their influence after projection through the higher-priority objectives. We therefore treat these objectives as \emph{free objectives}. They are not assigned a priority level, and their gradients are added directly to the accumulated projected gradient. This allows free objectives to regularize the motion without introducing additional nullspace constraints. Algorithm~\ref{alg:nsp} summarizes one iteration of the resulting prioritized optimization.

\paragraph{Warmup}
Prioritized optimization can converge slowly when the initial motion is far from a suitable solution. The nullspace projection restricts lower-priority updates according to the higher-priority objectives at every iteration, which can make it difficult to traverse large distances in the parameter space. We therefore begin with a short warmup period using a uniform weighted sum of objectives to move the motion closer to a suitable solution before applying the priority hierarchy.

\begin{algorithm}[!tb]
\caption{One iteration of prioritized optimization.}
\label{alg:nsp}
\begin{algorithmic}[1]
\Require $x$, prioritized objectives $\{(L^k,f^k)\}_{k=1}^K$, free objective $L_{\text{free}}$, damping $\lambda$, step size $\alpha$
\State $J^{<1} \gets [\,]$, \quad $g_{\text{total}} \gets 0$
\For{$k = 1,\ldots,K$}
  \State $g^k \gets \nabla_x L^k(x)$
    \State $\tilde g^k \gets
    \mathrm{Project}(g^k,J^{<k},\lambda)$
    \Comment{Eq.~\ref{eq:priority-per-frame-projection}}
  \State $g_{\text{total}} \gets g_{\text{total}} + \tilde{g}^k$ \Comment{accumulate projected gradient}
  \State $J^k \gets \partial f^k/\partial x$ \Comment{per-frame Jacobian}
  \State $J^{<k+1} \gets [J^{<k};\, J^k]$ \Comment{append rows to the stack}
\EndFor
\State $g_{\text{total}} \gets g_{\text{total}} + \nabla_x L_{\text{free}}(x)$ \Comment{free objectives}
\State $x \gets x - \alpha g_{\text{total}}$  \Comment{gradient update}\\
\Return $x$
\end{algorithmic}
\end{algorithm}

\section{Experimental Setup}
\label{sec:setup}
\method~is evaluated across retargeting settings ranging from single-character motion to multi-contact interactions involving multiple characters and objects. 
Our experiments primarily use the Unitree G1 as the target embodiment.
We evaluate retargeted motions in terms of both kinematic quality and their suitability as reference trajectories for reinforcement-learning policies.
Selected motions are additionally deployed on real humanoid robots.

\subsection{Motion Data \& Representation}
\label{sec:parametrization}
The motion of each target robot at frame $t$ is represented by its root position $r_t\in\R^3$, root orientation $\omega_t\in\R^3$, and joint configuration $\theta_t\in\R^{n_\theta}$. 
Each frame therefore contains $d=6+n_\theta$ parameters. 
For $N$ robots over $T$ frames, all robot motions are jointly optimized.
Before optimization, user-defined keypoint and joint correspondences are specified between each source character and target robot. 
Forward kinematics maps the motion parameters to the position $p_{t,i}\in\R^3$ and orientation $R_{t,i}\in SO(3)$ of keypoint $i$ at frame $t$, where $i$ indexes corresponding kinematic keypoints.

Single-character motion is evaluated on LAFAN1~\cite{harvey2020robust}, character--object interaction on the large-box category of OMOMO~\cite{li2023omomo}, and character--character interaction on the assistive subset of Inter-X~\cite{xu2024interx}, which features one character helping another rise from the ground.
We additionally evaluate complex multi-character motion using a 35-minute in-house dataset of collaborative object manipulation, including cooperative carrying and handoffs.

\begin{table}[!tb]
\caption{Error functions used in retargeting objective.}
\label{tab:objectives}
\centering
\footnotesize
\setlength{\tabcolsep}{4pt}
\begin{tabular}{@{}ll@{}}
\toprule
Objective & Error function $f$ \\
\midrule
\multicolumn{2}{@{}l}{\textit{Pose tracking}, for keypoint $i$} \\
\quad Position & $\hat p_{t,i} - s_\dagger\cdot\bar p_{t,i}$ \\
\quad Rotation & $\hat R_{t,i} \ominus \bar R_{t,i}$ \\
\addlinespace[2pt]
\multicolumn{2}{@{}l}{\textit{Interaction}, for keypoint pair $(i,j)$} \\
\quad Self & $(\hat p_{t,i} - \hat p_{t,j}) - s_{\mathrm{s}\rightarrow\mathrm{t}}(\bar p_{t,i} - \bar p_{t,j})$ \\
\quad Cross & $(\hat p^a_{t,i} - \hat p^b_{t,j}) - s_\dagger(\bar p^a_{t,i} - \bar p^b_{t,j})$ \\
\addlinespace[2pt]
\multicolumn{2}{@{}l}{\textit{Contact}, for active pair $(i,j)\in\mathcal{C}_t$} \\
\quad Object & $(\hat p_{t,i} - o_{t,j}) - s_\dagger(\bar{p}_{t,i} - o_{t,j})$ \\
\quad Character & $(\hat p^a_{t,i} - \hat p^b_{t,j}) - s_\dagger(\bar p^a_{t,i} - \bar p^b_{t,j})$ \\
\addlinespace[2pt]
\multicolumn{2}{@{}l}{\textit{Free objectives}} \\
\quad Temporal & $\theta_{t+1} - \theta_{t}$ \\
\quad Joint limits & $\max(0, \theta_t - \theta_t^{\max}) + \min(0, \theta_t - \theta_t^{\min})$ \\
\quad Ground & $\min(\hat p_{t,i}^{z},0)$ \\
\bottomrule
\end{tabular}
\vspace{-0.3cm}
\end{table}

\begin{table*}[!t]
\caption{Kinematic motion quality and tracking performance via RL.}
\label{tab:main}
\centering
\footnotesize
\setlength{\tabcolsep}{3pt}
\resizebox{\textwidth}{!}{%
\begin{tabular}{@{}llccccccccc@{}}
\toprule
 & & \multicolumn{2}{c}{\textbf{Penetration} $\downarrow$} & \textbf{Sliding} $\downarrow$ & \textbf{Floating} $\downarrow$ & \textbf{Jitter} $\downarrow$ & \textbf{Contact} $\uparrow$ & \multicolumn{3}{c}{\textbf{RL tracking}} \\
\cmidrule(lr){3-4} \cmidrule(lr){5-5} \cmidrule(lr){6-6} \cmidrule(lr){7-7} \cmidrule(lr){8-8} \cmidrule(l){9-11}
\textbf{Interaction Type} & \textbf{Method} & dur [\%] & depth [cm] & vel [cm/s] & height [cm] & [m/s$^3$] & [\%] & succ [\%] $\uparrow$ & err [cm] $\downarrow$ & obj [cm] $\downarrow$ \\
\midrule
\multirow{4}{*}{\shortstack[l]{\textbf{Single-character}\\\textit{(LAFAN1)}}}
 & GMR & $12.1^{\pm 19.1}$ & $10.7^{\pm 8.6}$ & $5.2^{\pm 1.8}$ & $2.1^{\pm 0.5}$ & $216.7^{\pm 71.4}$ & N/A & $90.0^{\pm 30.0}$ & $4.0^{\pm 6.1}$ & N/A \\
 & OmniRetarget & $\mathbf{1.5}^{\pm 3.8}$ & $\mathbf{0.7}^{\pm 1.0}$ & $9.4^{\pm 2.7}$ & $\mathbf{0.6}^{\pm 0.5}$ & $229.4^{\pm 64.0}$ & N/A & $90.0^{\pm 30.0}$ & $5.2^{\pm 11.3}$ & N/A \\
 & Ours & $12.2^{\pm 21.7}$ & $3.9^{\pm 3.8}$ & $\mathbf{3.5}^{\pm 1.1}$ & $1.4^{\pm 0.2}$ & $\mathbf{135.6}^{\pm 52.7}$ & N/A & $\mathbf{100.0}^{\pm 0.0}$ & $\mathbf{1.8}^{\pm 0.7}$ & N/A \\
\cdashline{2-11}
\addlinespace[2.5pt]
 & \textit{Source} & N/A & N/A & $1.2^{\pm 0.2}$ & $0.4^{\pm 0.4}$ & $126.5^{\pm 52.4}$ & N/A & N/A & N/A & N/A \\
\midrule
\multirow{5}{*}{\shortstack[l]{\textbf{Character--object}\\\textit{(OMOMO)}}}
 & PHC & $94.1^{\pm 9.9}$ & $9.3^{\pm 3.6}$ & $12.9^{\pm 3.2}$ & $\mathbf{0.2}^{\pm 0.3}$ & $\mathbf{76.3}^{\pm 14.6}$ & $63.2^{\pm 24.5}$ & $86.6^{\pm 34.0}$ & $7.4^{\pm 8.5}$ & $31.1^{\pm 70.3}$ \\
 & GMR & $65.4^{\pm 23.3}$ & $5.1^{\pm 3.4}$ & $\mathbf{1.7}^{\pm 0.5}$ & $\mathbf{0.1}^{\pm 0.5}$ & $94.7^{\pm 27.2}$ & $48.6^{\pm 22.4}$ & $79.8^{\pm 39.9}$ & $7.9^{\pm 9.8}$ & $38.7^{\pm 73.9}$ \\
 & OmniRetarget & $21.6^{\pm 22.9}$ & $2.7^{\pm 2.0}$ & $4.9^{\pm 0.8}$ & $0.3^{\pm 0.1}$ & $205.2^{\pm 47.6}$ & $86.4^{\pm 22.6}$ & $99.1^{\pm 2.9}$ & $3.5^{\pm 1.2}$ & $5.8^{\pm 6.3}$ \\
 & Ours & $\mathbf{13.4}^{\pm 16.4}$ & $\mathbf{1.8}^{\pm 1.9}$ & $3.7^{\pm 1.1}$ & $2.4^{\pm 0.1}$ & $108.0^{\pm 33.1}$ & $\mathbf{96.9}^{\pm 3.7}$ & $\mathbf{100.0}^{\pm 0.0}$ & $\mathbf{2.8}^{\pm 0.5}$ & $\mathbf{3.0}^{\pm 1.4}$ \\
\cdashline{2-11}
\addlinespace[2.5pt]
 & \textit{Source} & N/A & N/A & $1.2^{\pm 0.2}$ & $1.7^{\pm 0.1}$ & $54.5^{\pm 13.8}$ & N/A & N/A & N/A & N/A \\
\midrule
\multirow{5}{*}{\shortstack[l]{\textbf{Character--character}\\\textit{(Inter-X)}}}
 & PHC  & $79.0^{\pm 16.0}$ & $9.7^{\pm 3.1}$ & $8.8^{\pm 4.6}$ & $2.8^{\pm 3.3}$ & $\mathbf{84.0}^{\pm 23.6}$ & $35.2^{\pm 24.1}$ & $72.8^{\pm 43.2}$ & $8.7^{\pm 8.6}$ & N/A \\
 & GMR  & $67.8^{\pm 19.2}$ & $7.1^{\pm 2.5}$ & $3.9^{\pm 3.8}$ & $1.3^{\pm 2.2}$ & $147.2^{\pm 88.1}$ & $26.4^{\pm 19.4}$ & $57.2^{\pm 47.8}$ & $15.9^{\pm 13.5}$ & N/A \\
 & OmniRetarget  & $\mathbf{33.6}^{\pm 26.0}$ & $\mathbf{5.7}^{\pm 3.6}$ & $3.4^{\pm 0.7}$ & $\mathbf{0.2}^{\pm 0.2}$ & $136.7^{\pm 40.3}$ & $26.9^{\pm 21.6}$ & $66.1^{\pm 44.7}$ & $12.1^{\pm 11.9}$ & N/A \\
 & Ours & $55.7^{\pm 18.6}$ & $6.9^{\pm 2.0}$ & $\mathbf{2.4}^{\pm 0.8}$ & $1.8^{\pm 0.6}$ & $103.9^{\pm 47.6}$ & $\mathbf{78.8}^{\pm 13.7}$ & $\mathbf{93.6}^{\pm 24.0}$ & $\mathbf{4.5}^{\pm 5.8}$ & N/A \\
\cdashline{2-11}
\addlinespace[2.5pt]
 & \textit{Source} & N/A & N/A & $1.6^{\pm 0.3}$ & $2.3^{\pm 0.7}$ & $69.7^{\pm 40.8}$ & N/A & N/A & N/A & N/A \\
\midrule
\multirow{4}{*}{\shortstack[l]{\textbf{Multi-character + object}\\\textit{(in-house)}}}
 & GMR  & $39.3^{\pm 23.0}$ & $6.2^{\pm 5.1}$ & $4.4^{\pm 4.4}$ & $5.1^{\pm 1.5}$ & $102.3^{\pm 84.2}$ & $29.3^{\pm 22.1}$ & $24.1^{\pm 37.5}$ & $30.9^{\pm 15.2}$ & $137.1^{\pm 100.2}$ \\
 & OmniRetarget  & $\mathbf{10.2}^{\pm 15.5}$ & $\mathbf{2.0}^{\pm 3.7}$ & $2.7^{\pm 0.9}$ & $\mathbf{0.1}^{\pm 0.3}$ & $129.6^{\pm 36.5}$ & $26.5^{\pm 27.2}$ & $4.3^{\pm 13.0}$ & $36.8^{\pm 8.0}$ & $166.4^{\pm 73.7}$ \\
 & Ours & $26.4^{\pm 22.1}$ & $5.6^{\pm 3.7}$ & $\mathbf{1.7}^{\pm 0.9}$ & $3.2^{\pm 1.4}$ & $\mathbf{67.5}^{\pm 31.5}$ & $\mathbf{92.4}^{\pm 9.5}$ & $\mathbf{62.6}^{\pm 38.9}$ & $\mathbf{12.5}^{\pm 12.4}$ & $\mathbf{47.4}^{\pm 69.0}$ \\
\cdashline{2-11}
\addlinespace[2.5pt]
 & \textit{Source} & N/A & N/A & $0.6^{\pm 0.3}$ & $3.2^{\pm 1.1}$ & $35.4^{\pm 43.4}$ & N/A & N/A & N/A & N/A \\
\bottomrule
\end{tabular}}
\end{table*}

\subsection{Retargeting Configuration}
\label{sec:motion-objectives}

Table~\ref{tab:objectives} summarizes the error functions used in our experiments. Hats ($\hat{\cdot}$) and bars ($\bar{\cdot}$) denote quantities computed from the retargeted and source motions, respectively. 
For objectives involving pairs of keypoints, $j$ denotes the second keypoint, and superscripts $a$ and $b$ distinguish different characters. Interaction objectives preserve relative displacements between keypoint pairs selected using a per-frame Delaunay tetrahedralization of the source configuration~\cite{ho2010spatial}. Contact objectives are applied to pairs of bodies whose distance in the source motion falls below a threshold $c_{\mathrm{th}}$. Temporal smoothness, joint limits, and ground penetration are treated as free objectives throughout, as described in Section~\ref{sec:efficient}.

For single-character motions, the source motion is uniformly scaled to account for differences in the robot morphology using $s_{\mathrm{s}\rightarrow\mathrm{t}}=h_{\mathrm{tgt}}/h_{\mathrm{src}}$, where $h$ is measured from the toes to the shoulders. For motions involving multiple characters or objects, the source motion is instead scaled using a shared scene scale $s_\dagger$. 
For object interactions, we set $s_\dagger=1$ to preserve the geometry and trajectory of the object. For character--character interactions, one character is designated as the \emph{anchor}, and $s_\dagger$ is set to its source-to-target scale $s_{\mathrm{s}\rightarrow\mathrm{t}}$. Uniformly scaling all characters using this shared scale preserves their relative spatial relationships in the source motion.

A shared priority hierarchy is used across motions: (1) foot-position matching, (2) contact preservation with objects or other characters, when applicable, and (3) the remaining pose and interaction objectives. Foot-position matching preserves the stepping pattern of the source motion, while contact preservation enforces interaction contacts. The remaining pose and interaction objectives then use the available degrees of freedom to more closely match the source motion. While a general hierarchy is used in our experiments, specialized hierarchies better aligned with specific retargeting goals can also be used, as discussed further in Sec.~\ref{sec:discussion}.

\begin{figure}[!tb]
\centering
\includegraphics[width=1.0\columnwidth]{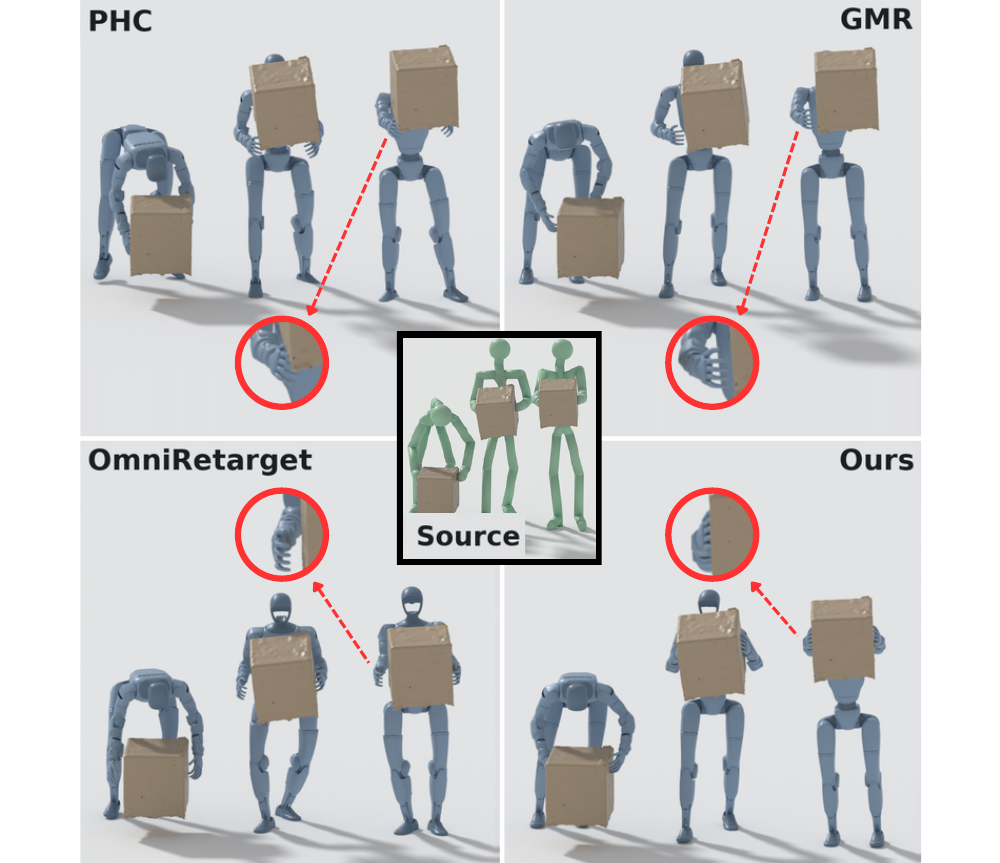}
\caption{Comparison of retargeting methods on a box-manipulation sequence. Baseline methods tend to distort the interactions or lose contact with the box, while \method~preserves the interactions more closely.}
\label{fig:qualitative-comparison}
\vspace{-0.3cm}
\end{figure}

\section{Results}
\label{sec:results}

\method~is compared against three widely used baselines for motion retargeting to humanoid robots: PHC~\cite{luo2023perpetual}, GMR~\cite{araujo2025retargeting}, and OmniRetarget~\cite{yang2025omniretarget}.
PHC performs trajectory-level keypoint fitting, GMR solves frame-wise inverse kinematics with a weighted-sum objective, and OmniRetarget uses constrained optimization based on interaction meshes.
For multi-character retargeting, baselines are applied independently for each character.

Kinematic motion quality is evaluated using metrics for penetration, foot sliding, foot floating, jitter, and contact preservation. 
\emph{Penetration} measures the duration and depth of intersections with the ground, objects, or other robots. 
\emph{Foot sliding} and \emph{foot floating} measure foot velocity and height during source stance phases, respectively. 
\emph{Jitter} is estimated from the third derivative of body positions, approximated using finite differences. 
\emph{Contact preservation} measures the fraction of source hand contacts reproduced after retargeting. 
Where applicable, the same metrics are reported for the source motion as a reference.

To assess suitability for RL training, we use retargeted motions as reference trajectories for motion-tracking controllers. 
Following DeepMimic~\cite{peng2018deepmimic}, we train a separate PPO policy~\cite{schulman2017proximal} for each motion on a simulated Unitree G1, using fixed hyperparameters across retargeting methods. 
Policies observe target poses and output joint PD targets; object-interaction motions additionally observe object position and target trajectory, while multi-robot motions use a single policy to control all robots. 
Across 50 motions, we report success rate, body tracking error, and object tracking error when applicable. S
uccess requires completing the reference motion without any robot or object deviating by more than $0.5$\ m. 
Tracking errors are computed over the full motion with early termination disabled.

\begin{table}[!tb]
\caption{Comparison against weighted-sum variants} 
\label{tab:noprio}
\centering
\scriptsize
\setlength{\tabcolsep}{2.5pt}
\resizebox{\columnwidth}{!}{
\begin{tabular}{@{}lcccccc@{}}
\toprule
& \multicolumn{2}{c}{\textbf{Penetration} $\downarrow$} & \textbf{Sliding} $\downarrow$ & \textbf{Floating} $\downarrow$ & \textbf{Jitter} $\downarrow$ & \textbf{Contact} $\uparrow$ \\
\cmidrule(lr){2-3} \cmidrule(lr){4-4} \cmidrule(lr){5-5} \cmidrule(lr){6-6} \cmidrule(l){7-7}
\textbf{Variant} & dur [\%] & depth [cm] & vel [cm/s] & height [cm] & [m/s$^3$] & [\%] \\
\midrule
\multicolumn{7}{@{}l}{\textbf{Single-character} \textit{(LAFAN1)}} \\
Ours & $\mathbf{12.2}^{\pm 21.7}$ & $\mathbf{3.9}^{\pm 3.8}$ & $3.5^{\pm 1.1}$ & $1.4^{\pm 0.2}$ & $\mathbf{135.6}^{\pm 52.7}$ & N/A \\
\quad WS-uniform & $12.4^{\pm 21.8}$ & $5.4^{\pm 5.0}$ & $3.3^{\pm 0.9}$ & $1.1^{\pm 0.2}$ & $140.8^{\pm 54.9}$ & N/A \\
\quad WS-$10\times$ & $13.1^{\pm 22.0}$ & $6.1^{\pm 5.3}$ & $\mathbf{2.9}^{\pm 0.8}$ & $\mathbf{0.8}^{\pm 0.2}$ & $141.4^{\pm 55.0}$ & N/A \\
\midrule
\multicolumn{7}{@{}l}{\textbf{Character--object} \textit{(OMOMO)}}  \\
Ours & $13.4^{\pm 16.4}$ & $\mathbf{1.8}^{\pm 1.9}$ & $3.7^{\pm 1.1}$ & $\mathbf{2.4}^{\pm 0.1}$ & $108.0^{\pm 33.1}$ & $\mathbf{96.9}^{\pm 3.7}$ \\
\quad WS-uniform & $\mathbf{12.4}^{\pm 15.0}$ & $2.2^{\pm 2.4}$ & $9.4^{\pm 2.5}$ & $9.5^{\pm 2.2}$ & $\mathbf{86.2}^{\pm 25.6}$ & $67.2^{\pm 33.1}$ \\
\quad WS-$10\times$ & $14.4^{\pm 17.2}$ & $2.7^{\pm 2.9}$ & $\mathbf{3.4}^{\pm 0.8}$ & $2.9^{\pm 0.6}$ & $98.5^{\pm 29.1}$ & $88.5^{\pm 12.0}$ \\
\midrule
\multicolumn{7}{@{}l}{\textbf{Character--character} \textit{(Inter-X)}}  \\
Ours & $55.7^{\pm 18.6}$ & $\mathbf{6.9}^{\pm 2.0}$ & $2.4^{\pm 0.8}$ & $1.8^{\pm 0.6}$ & $103.9^{\pm 47.6}$ & $\mathbf{78.8}^{\pm 13.7}$ \\
\quad WS-uniform & $54.9^{\pm 19.2}$ & $6.9^{\pm 2.1}$ & $3.4^{\pm 1.0}$ & $2.1^{\pm 0.9}$ & $92.3^{\pm 44.2}$ & $73.7^{\pm 14.1}$ \\
\quad WS-$10\times$ & $\mathbf{54.0}^{\pm 19.7}$ & $6.9^{\pm 2.2}$ & $\mathbf{2.2}^{\pm 0.5}$ & $\mathbf{1.7}^{\pm 0.6}$ & $\mathbf{91.9}^{\pm 33.3}$ & $75.0^{\pm 14.7}$ \\
\midrule
\multicolumn{7}{@{}l}{\textbf{Multi-character + object} \textit{(in-house)}}  \\
Ours & $\mathbf{26.4}^{\pm 22.1}$ & $\mathbf{5.6}^{\pm 3.7}$ & $\mathbf{1.7}^{\pm 0.9}$ & $\mathbf{3.2}^{\pm 1.4}$ & $67.5^{\pm 31.5}$ & $\mathbf{92.4}^{\pm 9.5}$ \\
\quad WS-uniform & $35.0^{\pm 23.4}$ & $5.8^{\pm 3.5}$ & $3.4^{\pm 1.7}$ & $14.0^{\pm 4.6}$ & $\mathbf{52.9}^{\pm 24.8}$ & $88.8^{\pm 12.4}$ \\
\quad WS-$10\times$ & $30.5^{\pm 24.2}$ & $5.8^{\pm 3.9}$ & $1.9^{\pm 1.0}$ & $5.1^{\pm 2.0}$ & $59.1^{\pm 28.2}$ & $71.7^{\pm 24.2}$ \\
\bottomrule
\end{tabular}}
\end{table}

\subsection{Comparisons}
Table~\ref{tab:main} shows that \method~better preserves motion properties assigned higher priorities in the hierarchy. 
It achieves the highest contact preservation across all interaction settings and generally reduces foot sliding relative to the baselines. 
Figure~\ref{fig:qualitative-comparison} highlights examples where the baseline methods fail to preserve the source hand--object contacts, while \method~more closely maintains the intended interactions. 
The retargeted motions also remain relatively smooth. 
\method~and PHC~\cite{luo2023perpetual} generally produce lower jitter than the per-frame optimization methods, consistent with both approaches optimizing motion jointly over time rather than retargeting each frame independently. 
Higher penetration and foot floating in some settings may reflect artifacts already present in the source motion.

Downstream tracking performance shows a similar trend.
Across all datasets, policies trained on retargeted motions by \method~achieve the highest success rates and lowest body tracking errors. 
The benefits are more evident for interaction-rich motions, where preserving contacts and coordination provides a more consistent reference motion. 
Nevertheless, collaborative object manipulation in the in-house dataset remains challenging, highlighting the difficulty of jointly controlling multiple interacting robots and objects.

\begin{figure}[!tb]
\centering
\includegraphics[width=\columnwidth]{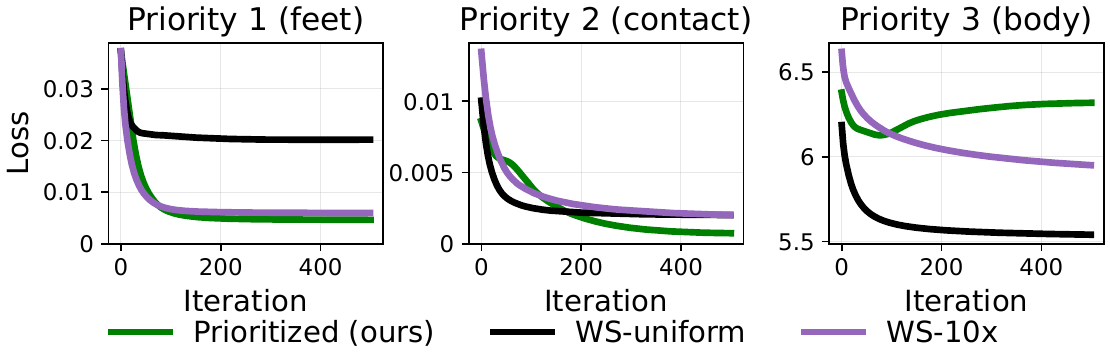} \\
\vspace{-0.3cm}
\caption{Loss curves for three objectives with different priority levels on a representative motion clip. Prioritized optimization reduces the higher-priority foot and contact losses at the expense of the lower-priority body objective, reflecting the relative importance of the objectives in the hierarchy.}
\label{fig:priority-losses}
\vspace{-0.3cm}
\end{figure}

\subsection{Effects of Prioritized Optimization}
\label{sec:prio-comparison}

To isolate the effect of prioritization, we compare \method~against two weighted-sum (WS) variants: \emph{WS-uniform}, which assigns equal weights to all objectives, and \emph{WS-$10\times$}, which uses a $10\times$ separation between objectives at successive priority levels. 
The latter approximates a stronger preference for higher-priority objectives using objective weighting alone.
Table~\ref{tab:noprio} shows similar performance for single-character locomotion, where objective conflicts are relatively limited and only two priority levels are present. 
Larger differences emerge in interaction settings, which introduce an additional contact-preservation priority level. 
Uniform weighting leads to worse foot and contact preservation, while increasing the weight separation recovers some of these properties but shifts the trade-off across priority levels. 
In contrast, \method~more consistently preserves higher-priority objectives without weight tuning.

Figure~\ref{fig:priority-losses} further illustrates this behavior for a representative motion sequence. 
\method~achieves the lowest losses at the first two priority levels, corresponding to foot matching and contact preservation, while incurring larger losses for the lower-priority body objectives. 
The weighted formulations reduce body loss further, but at the expense of higher-priority objectives.
This reflects the intended hierarchy: lower-priority objectives are improved only when doing so preserves higher-priority objectives.

\begin{table}[!tb]
\caption{Ablation of the weighted warmup and free objectives.}
\label{tab:ablation}
\centering
\scriptsize
\setlength{\tabcolsep}{2.5pt}
\resizebox{\columnwidth}{!}{%
\begin{tabular}{@{}lcccccc@{}}
\toprule
& \multicolumn{2}{c}{\textbf{Penetration} $\downarrow$} & \textbf{Sliding} $\downarrow$ & \textbf{Floating} $\downarrow$ & \textbf{Jitter} $\downarrow$ & \textbf{Contact} $\uparrow$ \\
\cmidrule(lr){2-3} \cmidrule(lr){4-4} \cmidrule(lr){5-5} \cmidrule(lr){6-6} \cmidrule(l){7-7}
\textbf{Variant} & dur [\%] & depth [cm] & vel [cm/s] & height [cm] & [m/s$^3$] & [\%] \\
\midrule
\multicolumn{7}{@{}l}{\textbf{Single-character} \textit{(LAFAN1)}} \\
Ours & $\mathbf{12.2}^{\pm 21.7}$ & $\mathbf{3.9}^{\pm 3.8}$ & $3.5^{\pm 1.1}$ & $1.4^{\pm 0.2}$ & $\mathbf{135.6}^{\pm 52.7}$ & N/A \\
\quad No warmup & $90.1^{\pm 15.3}$ & $14.1^{\pm 5.2}$ & $10.2^{\pm 2.1}$ & $\mathbf{0.0}^{\pm 0.1}$ & $168.1^{\pm 59.5}$ & N/A \\
\quad Free-last & $17.3^{\pm 23.0}$ & $6.9^{\pm 5.4}$ & $\mathbf{2.7}^{\pm 0.6}$ & $0.4^{\pm 0.2}$ & $160.4^{\pm 56.1}$ & N/A \\
\midrule
\multicolumn{7}{@{}l}{\textbf{Character--object} \textit{(OMOMO)}} \\
Ours & $13.4^{\pm 16.4}$ & $1.8^{\pm 1.9}$ & $3.7^{\pm 1.1}$ & $2.4^{\pm 0.1}$ & $108.0^{\pm 33.1}$ & $96.9^{\pm 3.7}$ \\
\quad No warmup & $\mathbf{12.1}^{\pm 16.1}$ & $2.0^{\pm 2.3}$ & $4.6^{\pm 1.7}$ & $3.0^{\pm 0.9}$ & $\mathbf{107.0}^{\pm 33.4}$ & $93.8^{\pm 6.6}$ \\
\quad Free-last & $14.4^{\pm 18.0}$ & $\mathbf{1.7}^{\pm 1.8}$ & $\mathbf{2.9}^{\pm 1.1}$ & $\mathbf{2.3}^{\pm 0.1}$ & $121.9^{\pm 39.6}$ & $\mathbf{99.3}^{\pm 1.2}$ \\
\midrule
\multicolumn{7}{@{}l}{\textbf{Character--character} \textit{(Inter-X)}} \\
Ours & $\mathbf{55.7}^{\pm 18.6}$ & $\mathbf{6.9}^{\pm 2.0}$ & $\mathbf{2.4}^{\pm 0.8}$ & $1.8^{\pm 0.6}$ & $\mathbf{103.9}^{\pm 47.6}$ & $78.8^{\pm 13.7}$ \\
\quad No warmup & $56.5^{\pm 19.2}$ & $7.0^{\pm 2.1}$ & $3.5^{\pm 1.6}$ & $1.9^{\pm 0.6}$ & $108.8^{\pm 50.9}$ & $70.3^{\pm 15.0}$ \\
\quad Free-last & $57.6^{\pm 19.2}$ & $7.0^{\pm 2.1}$ & $3.5^{\pm 1.5}$ & $\mathbf{1.6}^{\pm 0.7}$ & $155.0^{\pm 64.3}$ & $\mathbf{80.8}^{\pm 12.6}$ \\
\midrule
\multicolumn{7}{@{}l}{\textbf{Multi-character + object} \textit{(in-house)}}  \\
Ours & $\mathbf{26.4}^{\pm 22.1}$ & $\mathbf{5.6}^{\pm 3.7}$ & $\mathbf{1.7}^{\pm 0.9}$ & $\mathbf{3.2}^{\pm 1.4}$ & $\mathbf{67.5}^{\pm 31.5}$ & $\mathbf{92.4}^{\pm 9.5}$ \\
\quad No warmup & $35.1^{\pm 22.1}$ & $7.6^{\pm 3.8}$ & $4.0^{\pm 1.9}$ & $4.4^{\pm 1.9}$ & $72.9^{\pm 32.7}$ & $78.5^{\pm 14.0}$ \\
\quad Free-last & $26.6^{\pm 22.1}$ & $5.6^{\pm 3.7}$ & $2.3^{\pm 1.3}$ & $3.2^{\pm 1.4}$ & $88.6^{\pm 31.6}$ & $92.3^{\pm 9.6}$ \\
\bottomrule
\end{tabular}}
\end{table}

\begin{figure}[!tb]
\centering
\includegraphics[width=1.0\columnwidth]{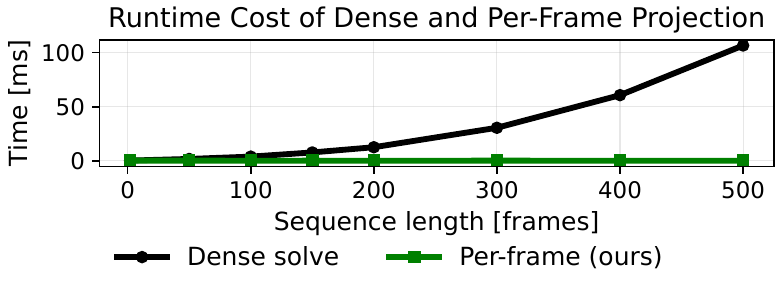} \\
\vspace{-0.4cm}
\caption{Runtime cost of forming the nullspace projection over different sequence lengths. The dense solve scales cubically, whereas our per-frame formulation remains nearly constant, largely dominated by GPU overhead.}
\label{fig:wall_clock}
\vspace{-0.4cm}
\end{figure}

\begin{figure*}[!t]
  \centering
  \includegraphics[width=\textwidth]{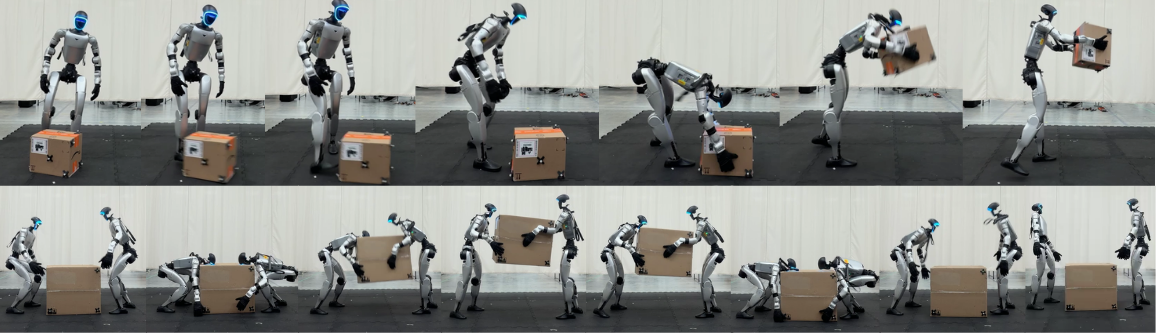}
    \caption{
    \textbf{Hardware deployment of retargeted box-manipulation motions on the Unitree G1.}
    \textbf{Top:} OMOMO manipulation sequence involving kicking and picking up a box.
    \textbf{Bottom:} Collaborative manipulation sequence collected in-house involving two robots picking up and placing down a large box.
    }
  \label{fig:robot_experiments}
  \vspace{-0.2cm}
\end{figure*}

\begin{figure}[!h]
\centering
\includegraphics[width=0.90\columnwidth]{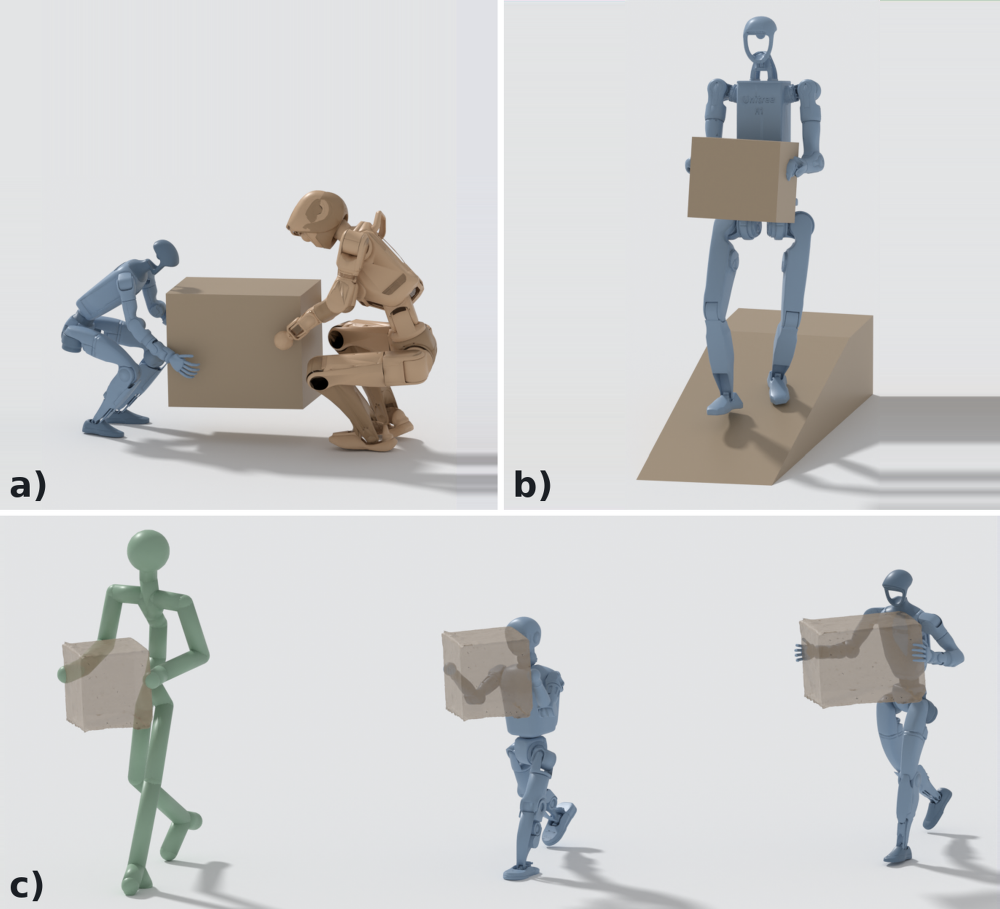}
\caption{\method~generalizes to different embodiments, scenes, and augmentations: \textbf{a)} a Unitree G1 and a Deep Robotics DR02 jointly carrying a box \textbf{b)} a Unitree H1-2 carrying a box down a ramp, and \textbf{c)} a box-carrying motion retargeted from the source (left) to a Booster T1 (center) and to a Unitree G1 (right), with the G1 holding a box enlarged by $1.5\times$.}
\label{fig:generality}
\vspace{-0.5cm}
\end{figure}

\subsection{Ablations}
\label{sec:ablations}

We ablate three key design choices: the weighted-sum warmup, free objectives, and per-frame projection. 
The effects of the warmup and free objectives are reported in Table~\ref{tab:ablation}. 
To isolate the effect of the warmup, we keep the number of prioritized optimization steps fixed when it is removed.
Without the warmup, performance changes little when the initialization is already close to a suitable solution, but degrades substantially when larger corrections are required.
This effect is most pronounced for LAFAN1 and the in-house motions, where the weighted-sum warmup helps move the initialization into a region where prioritized optimization can converge more effectively.

To evaluate the treatment of free objectives, we consider a \emph{Free-last} variant that places the free objectives inside the final priority level.
Projection through higher-priority objectives substantially limits their influence, particularly for temporal smoothness, resulting in increased jitter across settings.
Contact preservation improves slightly in some cases, but the gains are marginal.

Per-frame projection, introduced in Section~\ref{sec:efficient}, substantially reduces the computational cost of prioritized optimization. 
Figure~\ref{fig:wall_clock} compares its wall-clock cost against a dense projection across different motion sequence lengths on an NVIDIA RTX~5090. 
The cost of the dense formulation grows rapidly with sequence length, while the batched per-frame formulation remains nearly constant over the evaluated range, when runtime is dominated by GPU overhead.
We observe similar behavior for substantially longer sequences, including LAFAN1 motions exceeding 5000 frames.

\subsection{Generalization}
\label{sec:generality}

Figure~\ref{fig:generality} demonstrates the same prioritized formulation across a range of different embodiments, scenes, and object geometries. 
Collaborative manipulation is jointly retargeted to a Unitree G1 and DEEP Robotics DR02, which have substantially different morphologies.
A box manipulation motion is retargeted to a Unitree H1-2 robot traversing a ramp. 
The same OMOMO motion is also transferred to both a Booster T1 and a Unitree G1 using the same priority hierarchy. 
Scaling the manipulated object further demonstrates that the retargeted hand contacts adapt to changes in object geometry to preserve a high-priority contact objective, rather than strictly following the source motion. Demonstrations of policies deployed on real robots are shown in Figure~\ref{fig:robot_experiments} and the supplementary video.

\section{Discussion \& Future Work}
\label{sec:discussion}
In this work, we presented \method, a motion-retargeting method based on prioritized optimization.
Our experiments demonstrate its effectiveness as an alternative to balancing retargeting objectives through manually tuned loss weights.
\method~is not tied to a specific hierarchy, but instead provides a general mechanism for enforcing the relative importance of different objectives for a given motion.
In the current formulation, these priorities are specified manually.
Automatically determining or adapting the priority hierarchy is a promising direction for future work.

A key limitation of \method~ is that it primarily preserves the structure of the source motion, including any artifacts present in the original data.
Artifacts such as jitter, interpenetration, or inconsistencies in the captured environment can therefore propagate to the retargeted motion. 
Furthermore, the trajectories of objects being manipulated largely follow the original source motion, limiting the ability of the interaction to adapt when the object trajectory is poorly suited to the target embodiment.
Future work could address these limitations by incorporating learned motion priors, automatically adapting the priority hierarchy, and jointly optimizing robot and object trajectories.

\section*{Acknowledgments}
This work was supported in part by FieldAI, the Natural Sciences and Engineering Research Council of Canada (NSERC) through Discovery Grant RGPIN-2015-04843, and CIFAR though the CIFAR AI Chairs program.

\bibliographystyle{IEEEtran}
\bibliography{bibliography}
\end{document}